\documentclass[lettersize,journal]{IEEEtran}

\usepackage{color}
\usepackage{amsmath,amsfonts}
\usepackage{algorithmic}
\usepackage{algorithm}
\usepackage{array}
\usepackage{multicol}
\usepackage{multirow}
\usepackage[caption=false,font=normalsize,labelfont=sf,textfont=sf]{subfig}
\usepackage{textcomp}
\usepackage{stfloats}
\usepackage{url}
\usepackage{verbatim}
\usepackage{graphicx}
\usepackage{cite}
\usepackage{booktabs}

\begin{document}
\title{Lightweight Adaptive Feature De-drifting for Compressed Image Classification}

\author{Long~Peng*, Yang~Cao*, Member, IEEE,
        Yuejin~Sun, Student Member, IEEE and Yang~Wang† 
\thanks{

* Long Peng and Yang Cao contributed equally to this paper.

† Yang Wang is the corresponding author of this paper.

This work was supported by the Natural Science Foundation of China, No. 62206262. Yang Wang, Yuejun Sun, and Yang Cao are with the School of Information Science and Technology, University of Science and Technology
of China, Hefei 230026, China. (e-mail: ywang120@ustc.edu.cn, yjsun97@mail.ustc.edu.cn, forrest@ustc.edu.cn). Long Peng is with the Institute of Advanced Technology, University of Science and Technology of China, Hefei 230026, China. (e-mail: longp2001@mail.ustc.edu.cn).
}}

\markboth{IEEE Transactions on Multimedia}%
{Shell \MakeLowercase{\textit{et al.}}: A Sample Article Using IEEEtran.cls for IEEE Journals}

\maketitle
\begin{abstract}
JPEG is a widely used compression scheme to efficiently reduce the volume of the transmitted images at the expense of visual perception drop. The artifacts appear among blocks due to the information loss in the compression process, which not only affects the quality of images but also harms the subsequent high-level tasks in terms of feature drifting. High-level vision models trained on high-quality images will suffer performance degradation when dealing with compressed images, especially on mobile devices. In recent years, numerous learning-based JPEG artifacts removal methods have been proposed to handle visual artifacts. However, it is not an ideal choice to use these JPEG artifacts removal methods as a pre-processing for compressed image classification for the following reasons: 1) These methods are designed for human vision rather than high-level vision models. 2) These methods are not efficient enough to serve as a pre-processing on resource-constrained devices. To address these issues, this paper proposes a novel lightweight adaptive feature de-drifting module (AFD-Module) to boost the performance of pre-trained image classification models when facing compressed images. First, a Feature Drifting Estimation Network (FDE-Net) is devised to generate the spatial-wise Feature Drifting Map (FDM) in the DCT domain. Next, the estimated FDM is transmitted to the Feature Enhancement Network (FE-Net) to generate the mapping relationship between degraded features and corresponding high-quality features. Specially, a simple but effective RepConv block equipped with structural re-parameterization is utilized in FE-Net, which enriches feature representation in the training phase while keeping efficiency in the deployment phase. After training on limited compressed images, the AFD-Module can serve as a “plug-and-play” module for pre-trained classification models to improve their performance on compressed images. Experiments on images compressed once (\emph{i.e.} ImageNet-C) and multiple times demonstrate that our proposed AFD-Module can comprehensively improve the accuracy of the pre-trained classification models and significantly outperform the existing methods.
\end{abstract}

\section{Introduction}

\begin{figure}[tbp]  
    \centering
    
    \includegraphics[width=1.0\linewidth]{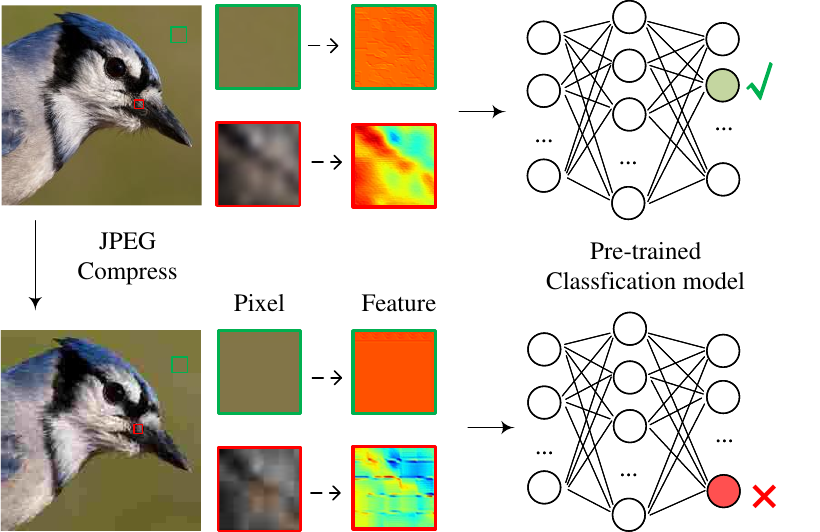}
    \caption{The degree of quality degradation varies greatly from region to region and has a strong correlation with the richness of details. The regions with more structure details suffer more high-frequency information loss, which leads to more heavy feature drifting problems for image recognition.}
\label{fig-prob1}
\end{figure}

\begin{figure*}[tbp]
\centering
\centerline{\includegraphics[width=1.0\linewidth]{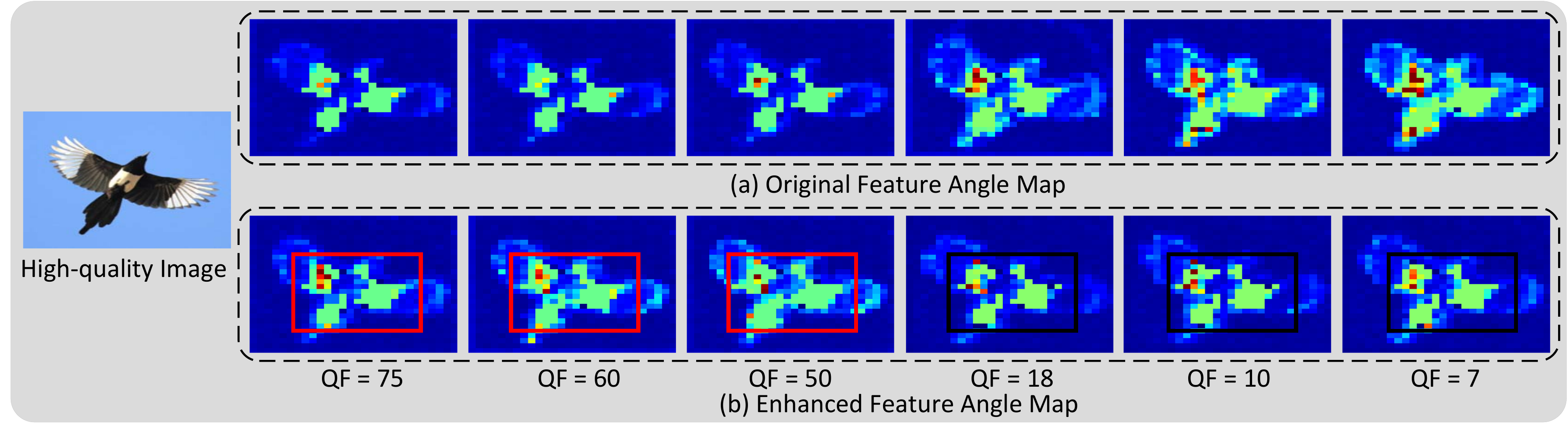}}
\caption{The artifacts induced by JPEG compression will lead to feature drifting for pre-trained models. We extract the features of high-quality blocks and compressed blocks from the first convolution layer of pre-trained RepVGG-A2. Then, we calculate the angle between the degraded features and the corresponding high-quality features for each pair of blocks, as shown in (a). Since JPEG compression applies DCT and quantization to each block individually, the correlation between adjacent blocks is ignored, which results in the spatially varied feature drifting. With the increase of compression degree, the inconsistency of feature drifting across regions increases gradually. (b) The feature angle map between high-quality features and enhanced features by MemNet \cite{tai2017memnet}. The MemNet can alleviate feature drifting to some extent for images with high compression ratio, as shown in the red box but introduce the opposite effect for images with low compression ratio, as shown in the black box.}
\label{fig1-Problem}
\end{figure*}

\begin{figure}[htbp]  
    \centering
    \includegraphics[width=1.0\linewidth]{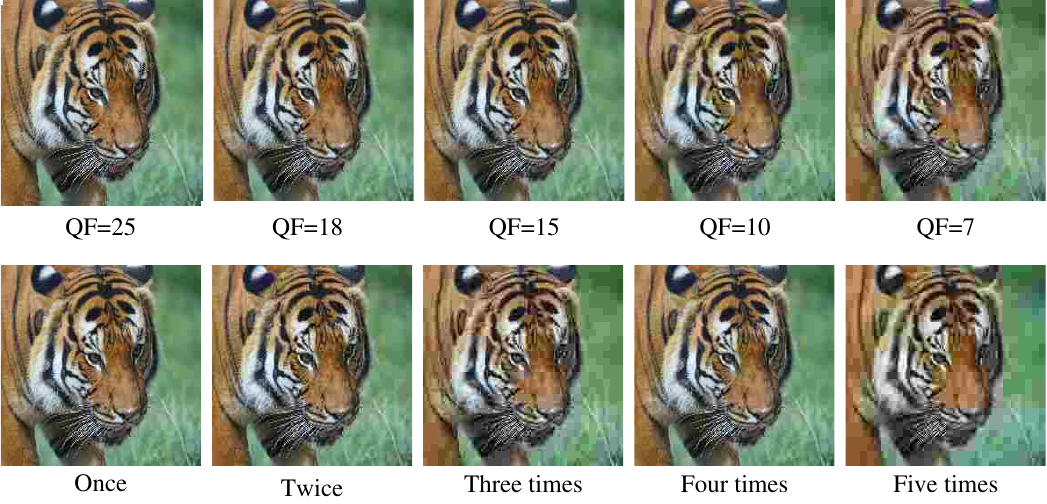}
    \caption{The degraded images from DIV2K-C \cite{agustsson2017ntire}. The first row represents different quality factors (QF) in a single compression. The second row represents the results in multiple compressions where QF is randomly selected from 25, 18, 15, 10, and 7.}
    \label{fig:example}
\end{figure}

\IEEEPARstart{W}ITH the development of imaging technology and the popularity of mobile multimedia, the spread of high-resolution images/videos is growing explosively. In order to save image/video transmission bandwidth and improve data transmission efficiency, lossy compression technology, such as JPEG, is widely embedded in mobile multimedia devices. However, the high-frequency signal of the image will be discarded in the lossy compression process, which will introduce artifacts into the image and reduce the image quality. As shown in Fig.~\ref{fig-prob1}, JPEG compression will result in multiple degrees of artifacts across the image. The regions with more tiny details will lose more high-frequency information compared to smoother regions. With the increase of image transmission times between mobile devices, the image will be continuously compressed multiple times, resulting in the artifact being gradually enlarged \cite{jiang2021towards}, as shown in Fig.\ref{fig:example}. These compression artifacts not only deteriorate the quality of visual perception but also destroy the image's structural and statistical properties, leading to feature fidelity degradation and performance degradation for high-level computer vision tasks. In this article, we focus on improving the recognition performance of JPEG-compressed images, which is the most widely used lossy compression technique in mobile devices. 

The reason for feature drifting is that lossy compression changes the inherent properties of images. The JPEG compression scheme divides an image into 8$\times$8 blocks and applies Discrete Cosine Transformation (DCT) and quantization to each block individually. However, due to the variations in structural and statistical properties across blocks, the lossy high-frequency components are also different for each block. And this leads to inconsistent feature drifting across blocks, resulting in a spatial-varied feature drifting for compressed images, as shown in Fig.~\ref{fig1-Problem} (a).

At present, using JPEG Artifact Removal (JAR) methods as a pre-processing to recover the compressed image and then performing image recognition is a typical solution \cite{liu2020comprehensive,ehrlich2020analyzing,galteri2017deep}. For images with a high compression ratio, these JAR methods could smooth the artifacts and improve the visual quality. This can suppress the feature drifting, as shown in the black rectangular box area in Fig.~\ref{fig1-Problem} (b), which can subsequently boost the image recognition performance. However, the JAR methods are not specially designed to recover the imperceptible tiny textures and tend to over-smooth the blocks, which is undesirable in low compression ratio\cite{liu2020comprehensive}. The smooth operation of the low compression ratio will result in the loss of details in the images. This will aggravate feature drifting, as shown in the red rectangular box area in Fig.~\ref{fig1-Problem} (b), which will further drop the recognition performance. Besides, the combination of JAR models and pre-trained models will heavily increase the overall parameters and computational costs, which makes it hard to employ on mobile devices. Other intuitive approaches for JPEG image recognition are data augmentation and domain adaptation. Using data augmentation to generate JPEG images and then retraining the model is straightforward. However, it will dramatically increase the amount of training data and computational cost and is difficult to apply in scenarios where training data is hard to available due to privacy protection, data storage, transmission costs, and computational burden \cite{yin2020dreaming,nayak2019zero}. Besides, the domain adaptation needs to collect the target data with the same semantic distribution as the source data, which is laborious. Furthermore, the models after domain adaption will inevitably suffer performance degradation in the source domain \cite{fang2022source,oza2023unsupervised}.

\begin{figure*}
\centering
\centerline{\includegraphics[width=1.0\linewidth]{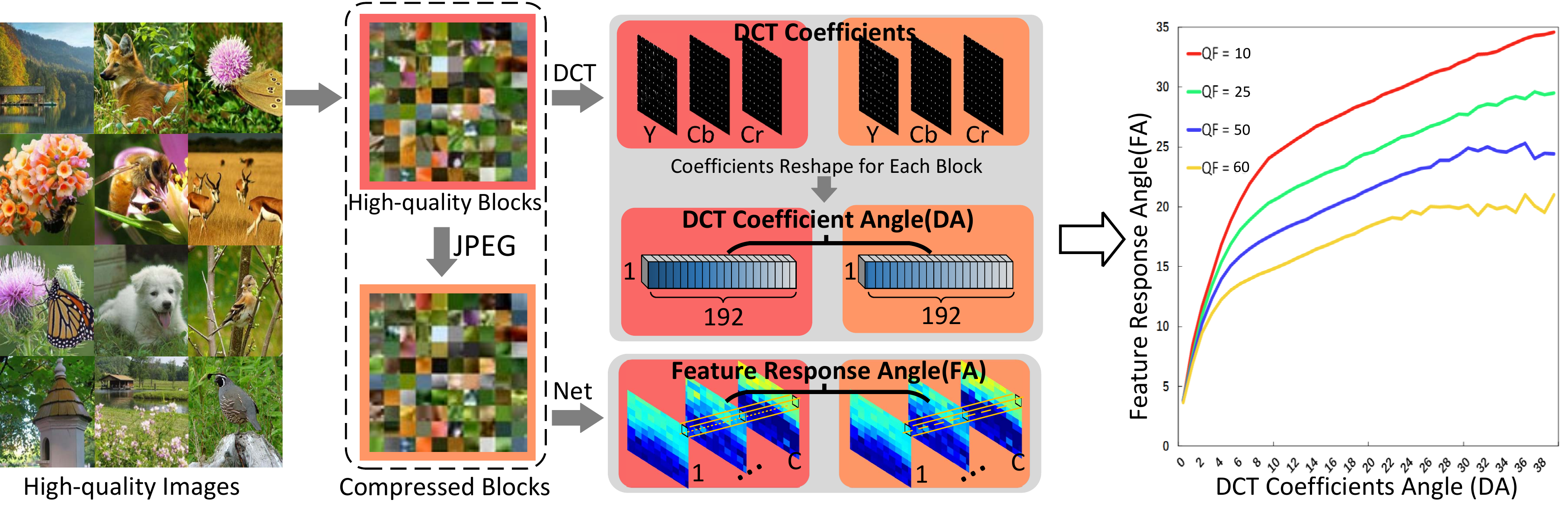}}
\caption{The illustration of statistical observations. We evaluate four JPEG compression ratios (QF = 10, 25, 50, and 60). The line with different colors represents the relationship between the DCT coefficients angle and feature response angle for images with different compression-ratio.}
\label{fig1-Motivation}
\end{figure*}

To address the above issues, this paper proposes to exploit the statistical property of JPEG images in the frequency domain to achieve an adaptive feature de-drifting for compressed image classification. The insight of our proposed method is derived based on the statistical observations shown in Fig.~\ref{fig1-Motivation} that is, for each block, the feature drifting has a strong correlation with the frequency statistical properties. Therefore, we propose to learn the mapping relationship between feature drifting and frequency distribution and leverage it as guidance for feature de-drifting. Note that the JPEG compression applies quantization to the frequency coefficients of each block individually and ignores the correlation between adjacent blocks. The compression result of each block is only related to the signal distribution within the local patch. The mapping relationship learned for one set of images can also be transferred to other sets of images.

Specifically, we propose an Adaptive Feature De-drifting Module (AFD-Module) to compensate for the varied spatial feature drifting for compressed image classification, which contains two sub-networks:  Feature Drifting Estimation Network (FDE-Net) and Feature Enhancement Network (FE-Net). Taking the rearranged DCT coefficients as input, FDE-Net is devised to estimate the Feature Drifting Map (FDM) for different blocks. Then, the FDM is transmitted to FE-Net to guide the feature de-drifting across blocks. Since JPEG compression is conducted in each $8\times8$ block that is irrelevant to the semantics of images, the statistical properties of blocks could be easily transferred~\cite{yang2010image} among different datasets. To further increase the inference speed of our method, we incorporate the reparameterize technology~\cite{ding2021repvgg} in our proposed RepConv Block, which can improve the model capacity by enriching the feature space without introducing extra parameters. After training on limited images without the supervision of semantic labels, the AFD-Module can be directly plugged into the existing classification networks to improve their performance on compressed images. For example, the proposed AFD-Module can be learned on 800 images from the DIV2K-C dataset and achieve the accuracy improvement of ImageNet-C \cite{hendrycks2019benchmarking} with 50,000 images in 1,000 classes. The contributions of this paper are three-fold:

(1) This paper proposes a novel Adaptive Feature De-drifting Module (AFD-Module), which perceives the feature drifting across the block and leverages it as the guidance map to guide the feature de-drifting. The statistical prior for feature drifting estimation is learned in the DCT domain, which makes the module transferable via limited training images.

(2) We adopt structural re-parameterization and introduce a multi-branch block in our proposed models to achieve a rich feature representation without introducing additional computational costs, which makes it feasible to be deployed on mobile devices.

(3) Experiments demonstrate that our proposed module achieves superior performance under both slight and severe JPEG compression conditions. Moreover, it is also promising to improve the recognition performance of images that have been compressed multiple times.

The rest of the paper is organized as follows. Firstly, we introduce relevant works. Then, the details of our proposed AFD-Module are represented in Sections \uppercase\expandafter{\romannumeral3}. The experimental results and ablation study are analyzed in Sections \uppercase\expandafter{\romannumeral4}. Finally, we give the conclusion in Sections \uppercase\expandafter{\romannumeral5}.

\begin{figure*}
\centering
\centerline{\includegraphics[width=1.0\linewidth]{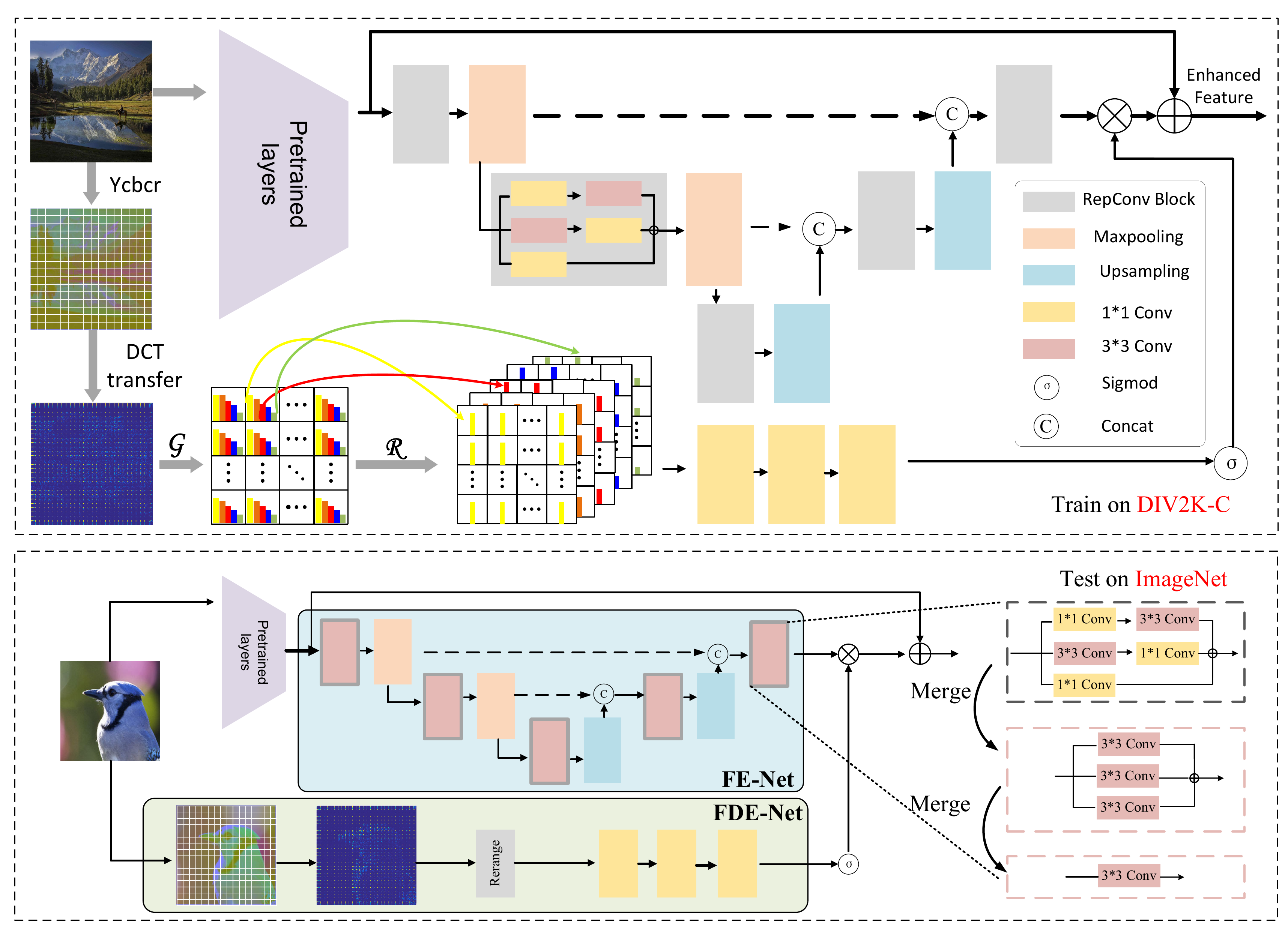}}
\caption{The architecture of our proposed Adaptive Feature De-drifting Module (AFD-Module). Our network contains two sub-networks: Feature Enhancement Network (FE-Net) and feature drifting Estimation Network (FDE-Net). The FE-Net is a lightweight U-Net-style network combined with the RepConv block. The FDE-Net is fed with the rearranged DCT coefficients, and the estimation map is obtained for spatial-wise degraded feature de-drifting. After training, the AFD-Module is plugged into the corresponding pre-trained model to boost the performance on compressed images.}
\label{fig1-TheFramework}
\end{figure*}
\section{Related Work}

In this section, we first review the current JPEG Artifacts Removal (JAR) methods. Then, we summary the works that focus on the combination of low-level tasks and high-level tasks. Finally, we introduce lightweight technologies, which aim at speeding up the inference time when deploying to mobile devices.

\subsection{JPEG Artifacts Removal Methods}
Traditional JAR methods try to remove image compression artifacts by manually designing various filters. For example, Foi~\textit{et al.}~\cite{foi2007pointwise} proposes a shape-adaptive DCT-based image filter for compression artifact removal. Norkin~\textit{et al.}~\cite{norkin2012hevc} proposes an in-loop filter to remove the artifacts in coding units. However, these above methods are time-consuming~\cite{fu2021model} and their representation capacity is constrained by the prior knowledge of filter design.

In recent years, numerous learning-based JAR methods have been proposed to remove the compression artifact and deliver a visually pleasing effect \cite{dong2015compression_ARCNN,zhang2017beyond_DNCNN,yoo2014post,galteri2019deep,li2017iterative,jiang2021towards,wang2022jpeg}. Typically, the ARCNN \cite{dong2015compression_ARCNN} first proposes a four-layer convolution neural network to learn the mapping from compressed images to clear images. To improve network performance and speed up the training process, residual learning and batch normalization are utilized in DnCNN \cite{zhang2017beyond_DNCNN} to remove image compression artifacts. The RED-Net \cite{mao2016image} introduces the deep residual encoding-decoding frameworks to fuse the features from different layers. The MemNet \cite{tai2017memnet} utilizes the dense connection framework and proposes a deep persistent memory network. To further utilize the characteristics of JPEG compression, some dual-domain methods for artifact removal are proposed. The Deep Dual-Domain (D$^3$) \cite{wang2016d3} combines both the prior knowledge in the JPEG compression scheme and the dual-domain sparse coding. The DMCNN \cite{zhang2018dmcnn} proposes a dual-domain encoder-decoder model to achieve better JAR performance. To further improve the details of the recovered image, RNAN \cite{rnan} employs non-local operators within convolution blocks to extract the cross-correlation in the image. To extract multi-scale features, DCSC \cite{dcsc} utilizes dilated convolutions to extract multi-scale features. Jiang \textit{et al.} first predicts the quality factor of the image and then embeds it in the JAR network \cite{jiang2021towards}. Wang \textit{et al.} uses contrastive learning to learn image compression quality representation and then proposes a blind JAR network \cite{wang2022jpeg}. These above JAR methods achieve superior performance for compression artifact removal and image quality improvement. However, these methods cannot guarantee that they will always be beneficial for downstream high-level vision tasks \cite{liu2020comprehensive}, especially when facing conditions with a high compression ratio. Besides, these methods are mainly devised to restore images that are compressed only once and rarely take multiple compression conditions into consideration.




\subsection{The Combination of Low- and High-level Models}
Degraded images will decrease the performance of high-level tasks (\emph{e.g.}~classification \cite{wang2020deep,krizhevsky2017imagenet,verma2018dct}, object detection \cite{chen2011detecting}). There are some works that propose the joint training strategy of low-level and high-level models to obtain task-specific enhancement models.

For example, Liu~\emph{et al.}~\cite{liu2017image} explores the connection between low-level vision and high-level vision tasks by cascading them together. Li~\emph{et al.}~\cite{li2017all} concatenates the image dehazing network and object detection network together to optimize them as a unified pipeline. Son~\emph{et al.}~\cite{son2020urie} proposes a training strategy that enables the enhancement model to be recognized successfully. There are other works that enhance the features to improve the performance under degraded conditions. For instance, Tan~\emph{et al.}~\cite{FSRGAN} proposes to super-resolution the features of small-size images in the deep representation space to enhance feature discrimination for image retrieval. Noh~\emph{et al.}~\cite{noh2019better} proposes a feature-level super-resolution approach to match the relative receptive fields of the input image so as to improve the performance of small object detection. Although the above methods are very effective, when semantic labels are not available, they are no longer available. However, these two works are not suitable for JPEG-compressed image recognition. Firstly, the above methods do not design a special removal method for the compression artifacts existing in the compressed features. Second, the above methods require the supervision of semantic signals. Further, DDP \cite{wang2020deep} proposes a deep degradation prior for low-contrast image classification, which does not need semantic labels for supervision. However, they do not take the difference of feature drifting across regions into consideration and cannot be directly used for compressed image classification.

\subsection{Lightweight Technology on Low-level Tasks}
With the development of deep learning, learning-based models show promising performance on various low-level tasks. However, they usually improve performance by increasing the number of network parameters and computational costs, which makes it difficult to deploy on resource-constrained mobile devices. To address these problems, various technologies have been proposed to pursue a trade-off between performance and efficiency. Some works\cite{sun2022lightweight,ahn2018fast, fu2019jpeg,wang2019progressive,wang2023decoupling,wang2023brightness} propose to design various lightweight architectures for low-level vision tasks (\emph{e.g.}, denoising, super-resolution, deraining). In these works, efficient UNet~\cite{unet} architecture and computational-friendly operators like Depthwise Convolution~\cite{howard2017mobilenets} are adopted to reduce the Floating Point Operations (FLOPs) and parameters. Other works attempt to reduce the computational costs by utilizing quantization~\cite{ma2019efficient, xin2020binarized}, pruning~\cite{li2020dhp}, and distillation~\cite{lee2020learning}.


Recently, structural re-parameterization~\cite{ding2021repvgg, ding2021diverse} has been proposed to learn multi-scale features to increase network representation capabilities in the training phase while maintaining a fast inference speed in the test phase through equivalent conversion. In this work, we introduce a multi-branch block in our feature enhancement module to extract multi-scale feature representations for better feature enhancement results. In the inference phase, the convolution block is transferred into a single convolution kernel with structural re-parameterization to speed up without performance drop.

\section{adaptive Feature De-drifting Module}
In this section, we first describe the statistical observation that inspires the design of the AFD-Module for feature de-drifting. Then, the details of the AFD-Module are introduced, which consists of FE-Net and FDE-Net. At last, the equivalent conversion of the RepConv block is defined in detail.

\subsection{The Statistical Observation}
The devised FDE-Net is inspired by the observation that the feature drifting has a strong correlation with the frequency statistical properties for each block. To illustrate the above observation, we perform a statistical experiment on high-quality and compressed images generated using the method in ImageNet-C \cite{hendrycks2019benchmarking}. As shown in Fig.~\ref{fig1-Motivation}, we first randomly sample 10,000 high-quality 8$\times$8 blocks from 200 images in ImageNet \cite{krizhevsky2017imagenet} and corresponding compressed blocks from the same positions. In this experiment, we verify four kinds of JPEG compression ratios (QF = 60, 50, 25, and 10). 

Secondly, the following terms for each pair of blocks are calculated: 1) We transfer high-quality blocks and corresponding compressed blocks into the DCT domain. Then, we calculate the vector angle of DCT coefficients (DA) defined by:
\begin{equation}
DA = \arccos \left( \frac{{d \cdot \tilde d}}{{\left\| d \right\| \cdot \left\| {\tilde d} \right\|}} \right),
\end{equation}
where $d$ and ${\tilde d}$ are 1$\times$192 DCT coefficients vector of high-quality block and corresponding compressed version. 2) We use the first convolution layer of pre-trained RepVGG-A2 to extract the feature responses of each block and evaluate the Feature drifting Angle (FA) defined by:
\begin{equation}
FA = \frac{1}{{64}}\sum\limits_{i = 1}^8 {\sum\limits_{j = 1}^8 {\arccos \left( {\frac{{{f_{i,j}} \cdot {{\tilde f}_{i,j}}}}{{\left\| {{f_{i,j}}} \right\| \cdot \left\| {{{\tilde f}_{i,j}}} \right\|}}} \right)} },
\label{fig:FA}
\end{equation}
where $f$ and ${\tilde f}$ are 1$\times$64 feature vector of high-quality block and corresponding compressed version. 

Finally, taking $DA$ as the horizontal axis, and $FA$ as the vertical axis, we draw the relationship curve between $DA$ and $FA$
as shown in Fig.~\ref{fig1-Motivation}. We can observe that: 1) The larger the DCT coefficients angle, the larger the corresponding feature drifting angle is. 2) The larger the degree of compression, the larger the drifting range of the feature is. We aim to leverage the frequency distribution within each compressed block to estimate the feature drifting map. To further facilitate the statistical modeling for feature drifting, we reaggregate the same frequency components across blocks into a single channel as shown in Fig.~\ref{fig1-TheFramework}. 
\subsection{The AFD-Module Architecture}
As shown in Fig.~\ref{fig1-TheFramework}, our AFD-Module contains two sub-networks: FE-Net and FDE-Net, which account for feature de-drifting and Feature Drifting Map (FDM) estimation, respectively.

\textbf{The FE-Net.} The input of our FE-Net is degraded features, which are extracted by the shallow layers of pre-trained models (\emph{e.g.}, layer1 in ResNet-50~\cite{resnet} and the first convolution layer in RepVGG-A2~\cite{ding2021repvgg}) and the corresponding high-quality features are used as ground truth. The shallow feature space mainly extracts the low-level features (\emph{e.g.}, textures, edges, and colors) \cite{asano2019critical}. Thus, the learned mapping relationship between degraded features and high-quality features on the one set (\emph{e.g.}, DIV2K-C) can be directly transferred to another set (\emph{e.g.}, ImageNet-C). In FE-Net, we adopt the residual learning formulation to train the mapping between degraded features and high-quality features. The Mean Squared Error (MSE) between the high-quality features and the estimated ones is adopted as the loss function to learn the trainable parameters of FE-Net. Fig.~\ref{fig1-TheFramework} illustrates the architecture of our proposed FE-Net, which is a lightweight U-Net style network that is composed of the RepConv block, which consists of three different routes to obtain multi-scale feature representation in the training phase. We adopt the sum operator for information fusion and apply classic ReLU~\cite{krizhevsky2017imagenet} as the non-linear activation. The estimated residual map is added to degraded features to recover the high-quality ones. To effectively alleviate the spatial-varied feature de-drifting, we introduce the FDM as guidance to achieve an adaptive feature de-drifting for different regions.

\textbf{The FDE-Net.}
The FDE-Net is devised to estimate Feature Drifting Map (FDM) by exploiting the frequency statistical properties within blocks. To facilitate the statistical modeling, we first transform the RGB color space to YCbCr and divide the image into 8$\times$8 blocks. Then, we transform each block into the DCT domain:
\begin{equation}
\begin{array}{l}
DCT(i,j) = \frac{1}{{\sqrt {2N} }}{\rm{C}}\left( i \right){\rm{C}}\left( j \right) \times \sum\limits_{x = 0}^{N - 1} {\sum\limits_{y = 0}^{N - 1} {Pixel(x,y)} }  \times \\
\\
\;\quad \;\quad \quad \;\;\;\;COS\left[ {\frac{{\left( {2x + 1} \right)i\pi }}{{2N}}} \right]COS\left[ {\frac{{\left( {2y + 1} \right)j\pi }}{{2N}}} \right]
\end{array}
\end{equation}

where ${ DCT }(i, j)$ is the 2D DCT frequency spectrum, and $C(x)=\frac{1}{\sqrt{2}} \text { if } x \text { is } 0, \text { else } 1 \text { if } x>0$. 
In FDE-Net, $N$ is equal to 8. Thus, a $3\times8\times8$ image block can be represented by 192 DCT coefficients. We gather the DCT coefficients on each block adaptively (\emph{i.e.} $\mathcal{G}$ in Fig.~\ref{fig1-TheFramework}). Then, all components of the same frequency are rearranged into one channel (\emph{i.e.} $\mathcal{R}$ in Fig.~\ref{fig1-TheFramework}). After that, an input 224$\times$224 RGB image can be represented as a $28\times28\times192$ DCT block, and the components with similar statistical properties within different image blocks are aggregated into a single channel. Taking the DCT block as input, our FDE-Net contains three stages, and each stage is composed of K$\times$K convolutional layers. In FDE-Net, we specifically set K equal to 1 to ensure that each convolution process only leverages the statistical properties of the current block and will not be influenced by adjacent blocks. The output FDM is transmitted to FE-Net to guide the feature enhancement. We up-sample the FDM by filling the corresponding feature area of each block with the same feature drifting estimation result to ensure scale consistency between feature maps in FE-Net and FDM.

\subsection{The RepConv Block}

The performance of learning-based methods always depends on the architecture and capacity of the model. Typically, the Inception\cite{szegedy2015going} models have promising performance because of their multi-branch architecture. The combination of the multiple paths with different scales can provide a very abundant feature space, which contributes to the large representational capacity of the model. However, such a multi-branch architecture inevitably increases computational costs and inference time, which limits a wider deployment on resource-constrained devices. A feasible solution is structural re-parameterization \cite{ding2021repvgg}, which can equivalently transfer the complex structure of models to lightweight models without performance drop when inferencing. During training, the multi-branch architecture ensures the learning capacity of models. After training, an equivalent transformation is performed to merge the multi-branch architecture into a single convolution to reduce computational costs and running time. 

Inspired by the Inception module\cite{szegedy2015going}, our RepConv block consists of three parallel branches. The first branch consists of serial 1$\times$1 and 3$\times$3 convolutions; the second branch consists of serial 3$\times$3 and 1$\times$1 convolutions; and the third branch only contains 1$\times$1 convolution. To meet the requirement of structural re-parameterization, we choose element addition operation to fuse multi-scale features. Then, we use Relu as the activation function, as shown in Fig \ref{fig1-TheFramework}. The details of the equivalent transformation can be summarized in the following three parts:

\subsubsection{Merge sequential convolutions} 

In the sequential combination structure of 1$\times$1 and 3$\times$3 convolution, we assume the kernel shapes of the 1$\times$1 and 3$\times$3 layers are D$\times$C$\times$1$\times$1 and E$\times$D$\times$3$\times$3, respectively, where D can be arbitrary. The sequential convolutions can be formulated as follows:
\begin{equation}
    W^{'} \otimes X_{in} = W^{3\times 3} \otimes\left (  W^{1\times 1} \otimes X_{in} \right ),
\end{equation}
where $ X_{in} $, $ \otimes $ denote input tensor and convolution operator, respectively.
$W^{1\times 1} \in  \mathbb {R}^{D\times C\times1\times1}$ and $W^{3\times 3} \in  \mathbb {R}^{E\times D\times3\times3}$ represent the weights of $1\times 1$, $3\times 3$ convolution kernel. C and D represent the number of input and output channels, respectively

As proved in \cite{ding2021diverse}, the sequential combination of  1$\times$1 and 3$\times$3 convolution can be merged to a single 3$\times$3 convolution operator. As $W^{1\times1}$ is 1$\times$1 convolution, which performs only channel-wise linear combination but no spatial aggregation, we can merge it into the 3$\times$3 convolution by linearly recombining the parameters in 3$\times$3 kernel. It is easy to verify that such a transformation can be accomplished by transpose convolution:

\begin{equation}
W^{'} = W^{3\times 3}\otimes TRANS\left ( W^{1\times 1} \right ). 
\end{equation}

where $TRANS$ is the tensor transposed from $W^{1\times 1}$. This equivalent procedure also works for the sequential combination structure of 3$\times$3 and 1$\times$1 convolution. More detailed proofs are referred to \cite{ding2021diverse}.

\subsubsection{Turn 1$\times$1 kernel to 3$\times$3 kernel}
Since 1$\times$1 convolution kernel cannot be directly added to the 3$\times$3 convolution kernel, we perform zero padding for 1$\times$1 convolution kernel to align the scale of 1$\times$1 kernel to 3$\times$3 kernel as:

\begin{equation}
     W_n^{3\times3} = Zeros(W^{1\times1}) 
\end{equation}
where the weights in 1$\times$1 kernel are used as the central weights of obtained $3\times3$ kernel $W_n$. In other words, within the $3\times3$ kernel, only the value of the central position is 1$\times$1 kernel, and the values of the rest of the positions are all 0.

\subsubsection{Merge parallel convolutions}
Through the above derivation, we can turn each branch in the RepConv block into a $3\times3$ kernel:

\begin{equation}
    W_{all} \otimes X_{in} =  W^{3\times3}_1 \otimes X_{in} + W^{3\times3}_2 \otimes X_{in} + W^{3\times3}_3 \otimes X_{in},
\end{equation}

Benefiting from the linearity of convolutions, the convolution kernels in parallel branches can be easily merged into a single convolution kernel~\cite{ding2021repvgg}, as in the following formula:

\begin{equation}
W_{all} = \sum_{i=1}^{3} W^{3\times3}_{i}.
\end{equation}

Finally, the RepConv block can be simplified to a single convolution kernel $W_{all}$, which provides multi-scale abundant features representation and keeps low inference computational costs.
\section{Experiments}

\subsection{Evaluation Metrics.}
The computational costs and memory requirements are critical factors in mobile device deployment. It is necessary for the model to achieve excellent classification performance while maintaining low computational costs and memory requirements. In this work, we take both efficiency (\emph{e.g.,}~FLOPs, Parameter, and Runtime) and effectiveness (\emph{e.g.,}~Accuracy) into consideration.

\subsubsection{Efficiency Evaluation}
In order to evaluate the performance on mobile devices, we test all the methods on the Redmi K30s mobile phone. The CPU and GPU of the tested mobile phone are the Snapdragon 865 and Adreno 640, respectively. We use the AI benchmark to evaluate the inference running time of different methods on the CPU and GPU. For each method, we calculate the average running time on the test set and then repeat it ten times to get the average result.

\subsubsection{Effectiveness Evaluation}
Following the setting of ImageNet-C~\cite{hendrycks2019benchmarking}, we test the Top-1 accuracy of the different classification models. The input size of both clean and compressed images is 224$\times$224.

\subsection{Experimental Setting}
We use the pre-trained ResNet-50~\cite{resnet}, and RepVGG-A2 \cite{ding2021repvgg} as the backbone. In our implementation, the proposed AFD-Module is trained for 15,000 iterations using Adam optimizer \cite{kingma2014adam} with an initial learning rate 1$\times10^{-4}$, batch size of 32. The learning rate decreases to 2$\times10^{-5}$ after 7,500 iterations. After training, we insert the pre-trained AFD-Module into existing classification networks to enhance the degraded features. We evaluate the proposed method on ImageNet-C \cite{hendrycks2019benchmarking} that has 50,000 images in 1,000 classes. Following the compressed image generation methods in ImageNet-C, we use the 800 high-quality images in DIV2k \cite{agustsson2017ntire} to synthesize five kinds of JPEG compressed images, randomly select 20,000 image patches with the resolution of 224$\times$224 for training and denote them as DIV2k-C.

\subsection{Ablation Study}

To evaluate the AFD-Module, we conduct experiments with several settings, including: 1) different depths of the pre-trained layers; 2) FDM estimation in DCT Domain; 3) shuffle DCT Coefficients; 4) Backbone of AFD-Module.

\subsubsection{Different depth of the pre-trained layers}
We conduct an ablation study on the depth of the pre-trained models. For ResNet-50, we denote the first convolution layer, the first ReLU layer, and the first BottleNeck in layer1 and layer1 with previous layers in ResNet-50 as D1–D4, respectively. For RepVGG-A2, we denote the stage0 layer, the first block in stage1, the stage1 layer, and the stage2 layer with previous layers in RepVGG-A2 as D1-D4. Deeper pre-trained layers will introduce more semantic information, which will influence the statistical modeling. As shown in Table \ref{tab:depth}, after enhancing the shallower features, the classification performance is better because the shallower features in the classification network contain more abundant low-level information. In the rest of the paper, we use the extracted features from the first ReLU layer of ResNet50 and the stage0 layer of RepVGG-A2 as the input of our AFD-Module to perform feature de-drifting for compressed image recognition.

\begin{table}[tbp]
\centering
  \caption{Classification performance comparison on different depths of the pre-trained layers.}
  \label{tab:depth}
\begin{tabular}{c||lllll}
\hline
\multicolumn{1}{l||}{}      & QF & D1    & D2    & D3    & D4    \\ \hline
\multirow{3}{*}{ResNet-50} & 25 & 65.82 & \textbf{67.10} & 66.25 & 64.18 \\
                           & 15 & 61.32 & \textbf{62.91} & 61.34 & 60.15 \\
                           & 7  & 47.59 & \textbf{49.02} & 48.32 & 47.32 \\ \hline
\multirow{3}{*}{RepVGG-A2} & 25 & \textbf{66.42} & 66.23 & 65.49 & 64.32 \\
                           & 15 & \textbf{62.35} & 61.98 & 58.23 & 57.89 \\
                           & 7  & \textbf{47.56} & 46.12 & 45.25 & 44.78 \\ \hline
\end{tabular}
\end{table}

\subsubsection{FDM estimation in DCT domain}
To demonstrate the benefits of discrete cosine transformation for feature drifting estimation, we compare the proposed FDE-Net (FDE-Net + rearranged DCT Coefficients) with its variants by replacing the DCT coefficients with image blocks. As shown in Table \ref{tab:FDM-PixleDomainFrequencyDomain}, compared with FDE-Net learned in the pixel domain (\emph{i.e.}, FE-Net + Image Block), the FDE-Net learned in the DCT domain can further improve the image recognition performance. The reason for this is that DCT aggregates the components with similar gradient characteristics in the image to form a more compact representation, which facilitates the statistical modeling for feature drifting estimation.

\begin{table}[tbp]
\centering
\caption{The benefits of FDM estimation in DCT domain.}
\resizebox{\linewidth}{!}{
\begin{tabular}{c||ccccc}
\hline
ResNet-50      & 25             & 18             & 10             & 7              \\ \hline
DirectTest     & 66.16          & 62.47          & 47.50          & 31.65          \\
FE-Net         & 66.25          & 63.34          & 53.45          & 45.48          \\
FE-Net + Image Block & 66.58          & 63.89          & 55.48          & 47.76          \\
FE-Net + Rearranged DCT Coefficients
    & \textbf{67.10} & \textbf{64.79} & \textbf{57.24} & \textbf{49.02} \\ \hline
\end{tabular}}
\label{tab:FDM-PixleDomainFrequencyDomain}%
\end{table}

\begin{table}[t]
\centering
\caption{Grouping the same frequency into a single channel can facilitate the FDM estimation and improve the compressed image classification accuracy.}
\footnotesize
\begin{tabular}{c||ccccc}
\hline
ResNet-50             & 25             & 18             & 10             & 7              \\ \hline
DirectTest            & 66.16          & 62.47          & 47.50          & 31.65          \\
Shuffled DCT Coefficients           & 64.87          & 63.21          & 54.59          & 46.28          \\
Rearranged DCT Coefficients & \textbf{67.10} & \textbf{64.79} & \textbf{57.24} & \textbf{49.02} \\ \hline
\end{tabular}
\label{tab:FDM-ShuffleDCT}%
\end{table}

\subsubsection{Shuffle DCT coefficients}
We also compare the difference between the random rearrangement of DCT coefficients (\emph{i.e.}, Shuffled DCT Coefficients) and the rearrangement of the same frequency coefficients into single channels (Rearranged DCT Coefficients). As shown in Table \ref{tab:FDM-ShuffleDCT}, the random arrangement of DCT coefficients can only improve the recognition performance on large compression ratio scenes. Further, the amplitude of improvement is much lower than grouping the same frequency into a single channel. This demonstrates the effectiveness of grouping the same frequency into a single channel for compressed image classification.
\subsubsection{Backbone of AFD-Module} To effectively enhance the details of compressed features and eliminate artifacts between blocks, we need to extract hidden correlations among blocks by utilizing the statistical and structural properties of adjacent blocks. To facilitate the capturing of statistical and structural properties, a typical and useful approach is to expand the receptive field of the network. Besides, the AFD-Module is a plug-and-play module that aims to assist the main pre-trained models to improve recognition under lossy compression conditions. Therefore, the AFD-Module should avoid introducing too much computation cost for pre-trained models. For the above considerations, we chose UNet as the backbone of the AFD-Module.

To further verify the rationality of using UNet as the backbone of AFD-Module, we compare the UNet backbone with other types of classic backbones. Specifically, \textbf{we keep the model parameters of other backbones almost the same as the UNet backbone} and replace UNet with the directly connected CNN backbone (DCNN), residual connection backbone (ResNet) \cite{resnet}, InceptionNet \cite{InceptionNet}, vision transformer (VIT) \cite{VIT}, and densely connected backbone DenseNet \cite{DenseNet} for training and evaluation. The results are shown in the Table \ref{table:Backbone}. Compared with other types of classic backbones, we can observe that UNet has the best performance under different compression rates.

\begin{table}[h]
\centering

\footnotesize
\caption{The ablation experiment on the backbone of AFD-Module. The best classification accuracy (\%) is marked as \textbf{bold}.}
    \resizebox{\linewidth}{!}{
\begin{tabular}{c|ccccccc}
\hline
                           & QF   & DCNN  & ResNet & Inception & VIT   & DenseNet & UNet \\ \hline
\multirow{6}{*}{{\rotatebox{270}{ResNet-50}}} & 25       & 65.87 & 66.20  & {66.87}       & 65.91 & 66.46    & \textbf{67.10}    \\
                           & 18       & 63.10 & 63.62  & {63.93}        & 63.16 & 63.86    & \textbf{64.79}      \\
                           & 15       & 60.26 & 61.09  & {61.95}        & 60.22 & 61.32    & \textbf{62.91}     \\
                           & 10        & 55.41 & 56.26  & {56.84}        & 55.99 & 56.64    & \textbf{57.24}    \\
                           & 7        & 47.64 & 47.92  & {48.19}        & 47.63 & 47.89    & \textbf{49.02}    \\
                           & Avg      & 58.46 & 59.02  & {59.55}       & 58.58 & 59.23    & \textbf{60.21}      \\ \hline
\end{tabular}
}
\label{table:Backbone}
\end{table}

\subsection{Quantitative Comparisons}
In this section, we compare our AFD-Module with seven representative JAR methods: ARCNN \cite{dong2015compression_ARCNN}, DnCNN \cite{zhang2017beyond_DNCNN}, RNAN \cite{rnan}, RDN \cite{zhang2019residual}, MIRNet \cite{mirnet}, FBCNN \cite{jiang2021towards}, JARCAL \cite{wang2022jpeg},  GRL \cite{GRL}, PANet \cite{PANet} and EARN \cite{EARN}. We train our AFD-Module on the DIV2k-C dataset and evaluate the performance under five JPEG compression conditions, respectively. For a fair comparison, the above JAR methods are also retrained on the DIV2k-C dataset. Apart from the above JAR methods, we also compare our method with DDP~\cite{wang2020deep} designed for feature enhancement for low-quality image classification, and the task-driven method URIE ~\cite{son2020urie}.

As shown in Table \ref{tab:quan on differeft factor}, the performance of pre-trained ResNet-50 drops from 76.14$\%$ to 31.65$\%$ when fed heavily compressed images (QF=7), which demonstrates that image classification on JPEG compressed images is very challenging. Although existing JAR methods can improve performance in most heavy compression conditions (\emph{e.g.} QF=7), they damage classification performance under slight compression conditions. For instance, the accuracy of ResNet-50 decreased to 66.16$\%$ when tested directly on compressed images of QF=25. The cascade combination of the JAR model and classification model cannot improve the performance and even drop the accuracy to 61.69$\%$ for ARCNN. Even worse, some JAR methods will degrade JPEG image recognition performance, especially in slight compression scenarios. For example, when QF is 25, we can observe that most of the JAR methods reduce the recognition performance of Resnet-50 and RepVGG-A2. The reason for this is that JAR methods are mainly devised to enhance the quality of compressed images and obtain visually pleasing enhancement results, but the information in the enhanced images cannot guarantee that it meets the featured distribution required by the recognition network. Compared with JAR methods, our method can improve the performance of ResNet-50 and RepVGG-A2 under both heavy and slight compression-ratio conditions, demonstrating that our method can learn a robustness mapping for degraded features de-drifting. Further, since the estimated FDM can effectively guide the feature enhancement, our method outperforms the existing methods by a large margin, especially for images with a heavy compression ratio.

In the real scene, images are often compressed multiple times through different digital media, so we also conduct multiple compression experiments. We perform multiple compressions on the dataset, and the quality factor of each compression is randomly chosen from 25, 18, 15, 10, and 7. In this paper, the maximum number of compressions is set to 5. As shown in Table \ref{tab:quan on compress times}, the performances of the pre-trained model decrease significantly with the increase of compression times. Compared with JAR methods, our proposed method achieves better performance, especially in the five compression times, which demonstrates that our method can better enhance degraded features transmitted by digital media.

\label{subsec:Comparison}

\begin{table*}[h!]
\centering
\caption{The image classification performance for quality factor 25, 18, 15, 10, 7 on ImageNet dataset. Avg means the average classification performance of different QF. The best classification accuracy (\%) is marked as \textbf{bold}}
  \label{tab:quan on differeft factor}
    \resizebox{\textwidth}{!}{
    
\begin{tabular}{c|cccccccccccccc}
\hline
                           & QF  & DirectTest & ARCNN & DnCNN & RNAN  & RDN   & MIRNet & FBCNN & JARCAL & PANet & EARN  & GRL   & DDP   & Ours           \\ \hline
\multirow{6}{*}{{\rotatebox{270}{ResNet-50}}} & 25  & 66.16      & 61.69 & 64.74 & 64.77 & 64.52 & 64.03  & 63.27 & 64.41  & 63.16 & 62.75 & 64.29 & 66.45 & \textbf{67.10} \\
                           & 18  & 62.47      & 60.35 & 61.89 & 62.25 & 63.15 & 62.79  & 61.04 & 62.04  & 61.67 & 60.17 & 62.17 & 63.68 & \textbf{64.79} \\
                           & 15  & 59.28      & 58.76 & 60.25 & 60.15 & 59.68 & 59.89  & 55.90 & 59.86  & 56.23 & 55.38 & 59.94 & 61.62 & \textbf{62.91} \\
                           & 10  & 47.50      & 53.10 & 53.32 & 53.59 & 52.25 & 51.78  & 47.05 & 53.16  & 47.43 & 47.84 & 53.76 & 55.37 & \textbf{57.24} \\
                           & 7   & 31.65      & 44.64 & 43.19 & 44.10 & 43.15 & 42.95  & 39.23 & 45.62  & 40.03 & 38.81 & 44.17 & 47.64 & \textbf{49.02} \\
                           & Avg & 53.41      & 55.71 & 56.68 & 56.97 & 56.55 & 56.29  & 53.30 & 57.02  & 53.71 & 52.99 & 56.87 & 58.95 & \textbf{60.21} \\ \hline
\multirow{6}{*}{\rotatebox{270}{RepVGG-A2}} & 25  & 65.59      & 61.99 & 63.78 & 64.13 & 64.57 & 63.98  & 62.28 & 64.82  & 63.23 & 62.27 & 64.24 & 65.74 & \textbf{66.42} \\
                           & 18  & 61.55      & 60.26 & 61.47 & 60.25 & 60.35 & 58.27  & 59.39 & 61.42  & 60.47 & 59.73 & 61.53 & 62.68 & \textbf{63.25} \\
                           & 15  & 58.24      & 58.52 & 59.46 & 58.16 & 59.78 & 59.45  & 57.47 & 59.43  & 59.08 & 58.30 & 59.47 & 60.28 & \textbf{62.35} \\
                           & 10  & 47.71      & 53.21 & 53.63 & 54.56 & 53.22 & 52.32  & 49.41 & 53.74  & 52.44 & 52.03 & 53.58 & 54.41 & \textbf{56.31} \\
                           & 7   & 34.20      & 44.30 & 44.83 & 43.65 & 42.98 & 41.26  & 42.30 & 44.75  & 43.42 & 42.60 & 44.70 & 46.80 & \textbf{47.56} \\
                           & Avg & 53.46      & 55.66 & 56.63 & 56.15 & 56.18 & 55.06  & 54.17 & 56.83  & 55.73 & 54.98 & 56.70 & 57.98 & \textbf{59.18} \\ \hline
\end{tabular}

}
\end{table*}

\begin{table*}[h!]
 \centering
  \caption{The image classification performance for multiple compression on ImageNet dataset. The quality factor in each compression is randomly selected from 25,18,15,10,7. CT and Avg mean compressed time and the average classification performance of different CT.}
  \label{tab:quan on compress times}
  \resizebox{\textwidth}{!}{
  
\begin{tabular}{c|cccccccccccccc}
\hline
                           & CT  & DirectTest & ARCNN & DnCNN & RNAN  & RDN   & MIRNet & FBCNN & JARCAL & PANet & EARN  & GRL   & DDP   & Ours           \\ \hline
\multirow{6}{*}{{\rotatebox{270}{ResNet-50}}} & 1   & 53.33      & 55.72 & 56.41 & 56.52 & 56.35 & 56.14  & 53.60 & 55.60  & 53.97 & 52.30 & 54.93 & 58.56 & \textbf{59.58} \\
                           & 2   & 45.44      & 52.02 & 52.13 & 51.58 & 51.14 & 49.68  & 49.40 & 50.17  & 49.50 & 48.96 & 49.83 & 52.69 & \textbf{54.85} \\
                           & 3   & 40.31      & 47.25 & 46.95 & 48.32 & 47.33 & 45.36  & 45.34 & 47.88  & 45.95 & 45.65 & 47.24 & 49.38 & \textbf{51.70} \\
                           & 4   & 36.59      & 44.26 & 45.37 & 46.32 & 45.08 & 42.10  & 41.90 & 44.67  & 42.59 & 41.06 & 44.79 & 47.74 & \textbf{49.41} \\
                           & 5   & 34.19      & 43.98 & 44.25 & 44.17 & 43.31 & 39.63  & 39.96 & 41.73  & 40.47 & 39.61 & 41.81 & 44.19 & \textbf{47.51} \\
                           & Avg & 41.97      & 48.65 & 49.02 & 49.38 & 48.64 & 46.58  & 46.04 & 48.01  & 46.50 & 45.52 & 47.72 & 50.51 & \textbf{52.61} \\ \hline
\multirow{6}{*}{\rotatebox{270}{RepVGG-A2}} & 1   & 53.44      & 55.72 & 55.14 & 55.23 & 55.94 & 54.15  & 53.43 & 55.17  & 53.93 & 52.19 & 55.16 & 56.65 & \textbf{58.16} \\
                           & 2   & 46.16      & 51.04 & 52.02 & 51.88 & 50.49 & 48.79  & 49.86 & 50.75  & 49.57 & 48.57 & 49.92 & 51.89 & \textbf{53.57} \\
                           & 3   & 41.76      & 46.72 & 47.25 & 48.32 & 46.87 & 45.54  & 45.08 & 47.76  & 46.31 & 45.95 & 47.27 & 49.89 & \textbf{51.03} \\
                           & 4   & 38.68      & 44.61 & 46.26 & 46.59 & 44.37 & 42.61  & 42.73 & 46.70  & 45.81 & 45.18 & 46.61 & 46.82 & \textbf{48.89} \\
                           & 5   & 36.19      & 43.85 & 43.98 & 44.15 & 42.46 & 40.49  & 42.05 & 44.12  & 43.40 & 42.70 & 44.02 & 46.25 & \textbf{47.23} \\
                           & Avg & 43.25      & 48.39 & 48.93 & 49.23 & 48.03 & 46.32  & 46.63 & 48.90  & 47.80 & 46.92 & 48.60 & 50.30 & \textbf{51.78} \\ \hline
\end{tabular}

}
\end{table*}

\begin{figure*}[tbp]
\centering
\centerline{\includegraphics[width=0.95\linewidth]{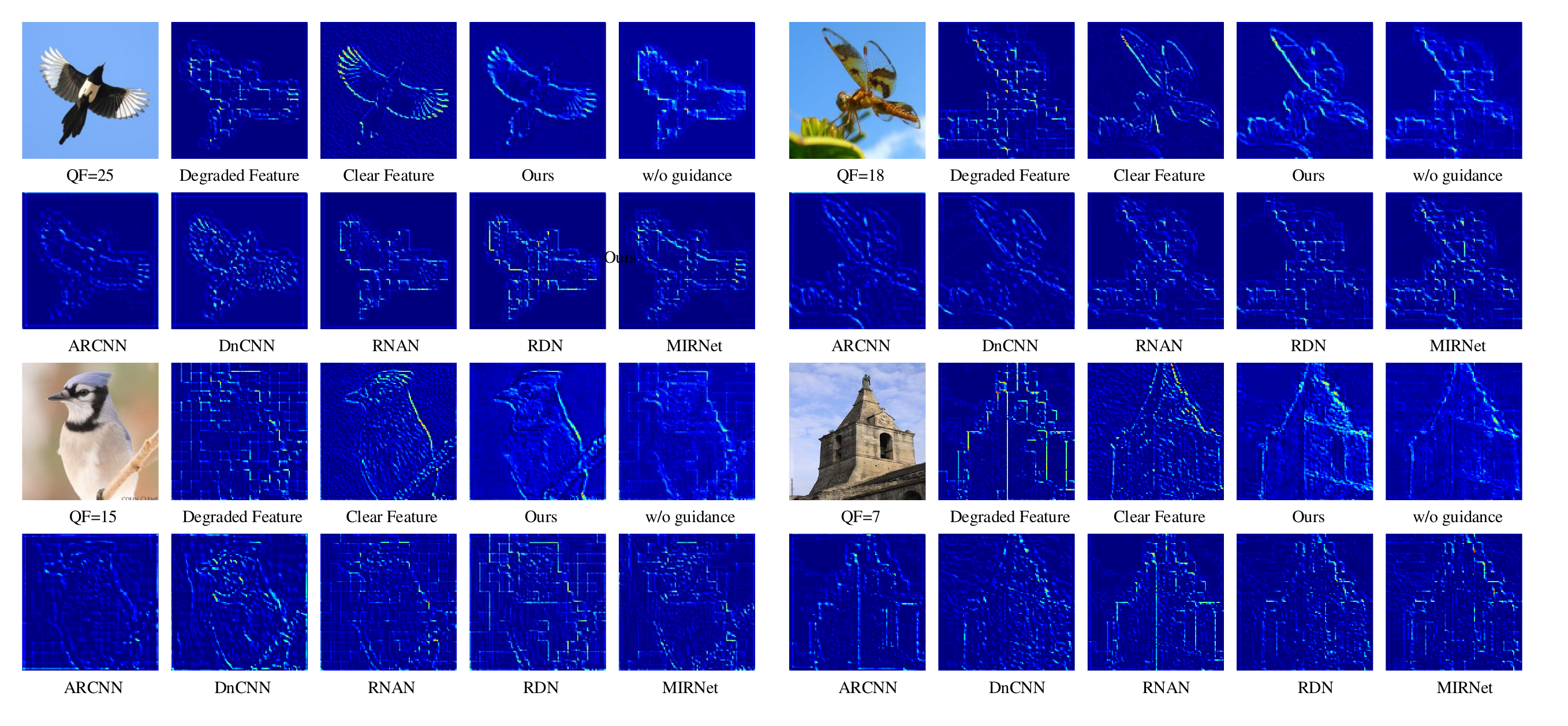}}
\caption{The feature maps of compressed images (QF = 25, 18, 15, 7) and the feature enhancement results using different compression artifact removal methods. The test images are randomly sampled from ImageNet-C. The feature maps are extracted from the ``first relu'' layer of ResNet-50 pre-trained on ImageNet. With the increase of compression degree, our method enables to recover the structural information while eliminating the spatial varied feature drifting.}
\label{fig1-CompareWithOtherMethods}
\end{figure*}

\begin{figure*}[h]
\begin{center}
\includegraphics[width=0.95\linewidth]{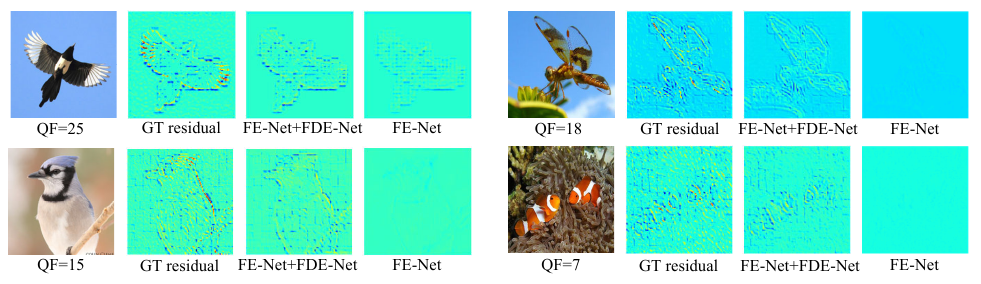}
\end{center}
    \caption{The comparison of residual maps estimated by FE-Net with or without the guidance of FDE-Net. Under the guidance of FDE-Net, FE-Net can accurately estimate the feature drifting under various compression conditions. Without the guidance of FDE-Net, FE-Net can only estimate the feature drifting on slight compression condition,\emph{i.e.} QF = 25.}
\label{fig:fig-ResidualComparison}
\end{figure*}

\begin{figure*}[h]
\vspace{-5mm}
\centering
\centerline{\includegraphics[width=0.95\linewidth]{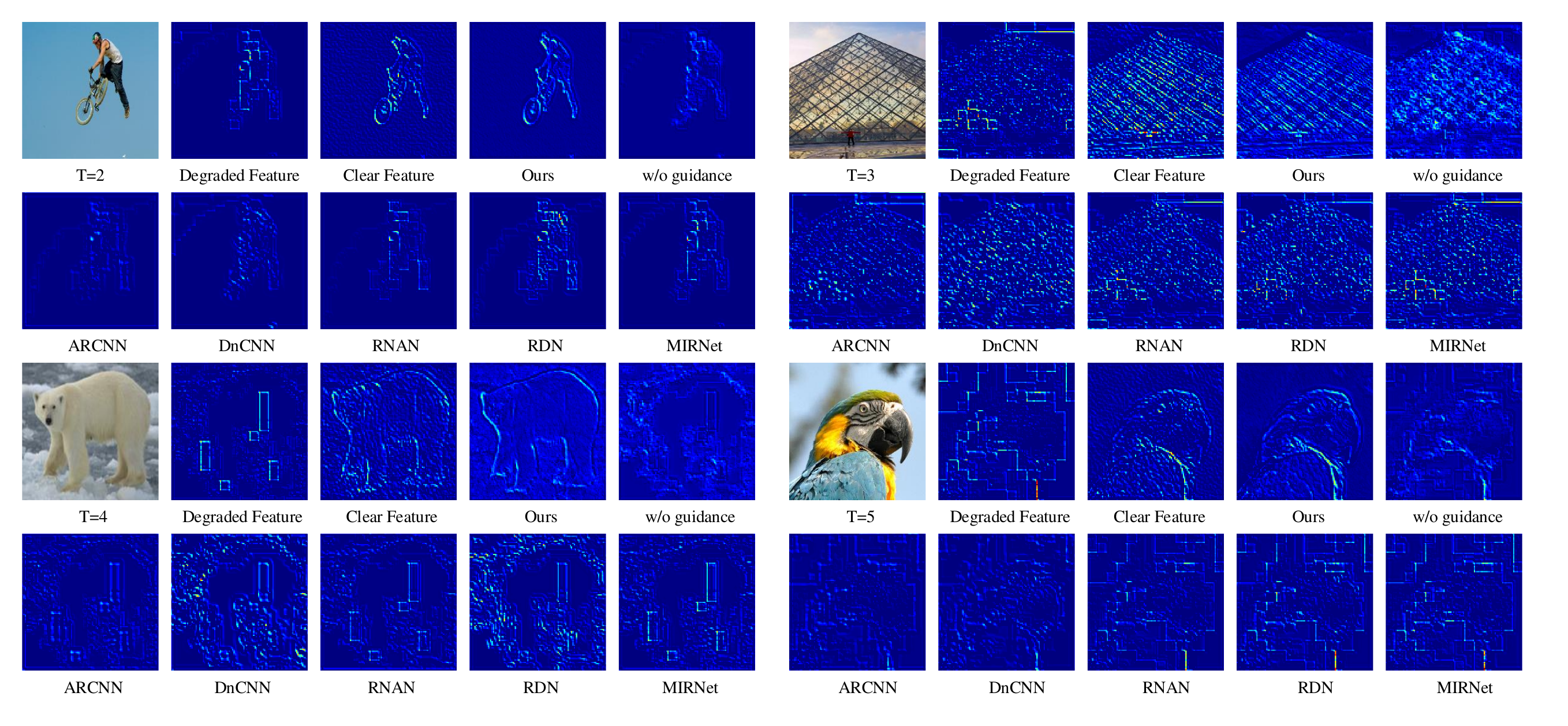}}
\vspace{-5mm}
\caption{The feature visualizations of multi-compressed images (compressed time T = 2, 3, 4, and 5) and the feature enhancement results using different compression artifact removal methods. The test images are randomly sampled from ImageNet-C. The feature maps are extracted from the ``first relu'' layer of ResNet-50 pre-trained on ImageNet. With the increase of compression times, our method enables to recover the structural information while eliminating the spatial varied feature drifting.}
\label{fig:fig1-multi-compressed}
\vspace{-5mm}
\end{figure*}

\begin{table}[h]
 \centering
  \caption{Comparison of inference time and computational costs on the mobile platform.}
  \label{tab:inf on mobile}
  
\begin{tabular}{c|cccc}
\toprule
\multicolumn{1}{l|}{} & \multicolumn{1}{l}{Params(M)} & \multicolumn{1}{l}{FLOPs(G)} & \multicolumn{1}{l}{CPU(ms)} & \multicolumn{1}{l}{GPU(ms)} \\ \hline
ResNet-50\cite{resnet}             & 25.5                          & 4.12                         & 132                         & 136                         \\
RepVGG-A2\cite{ding2021repvgg}             & 25.6                          & 5.12                         & 122                         & 125                         \\ \hline
ARCNN\cite{dong2015compression_ARCNN}                 & 0.12                          & 5.91                         & 185                         & 178                         \\
DnCNN\cite{zhang2017beyond_DNCNN}                 & 0.56                          & 28.07                        & 625                         & 1132                        \\
RNAN\cite{zhang2019residual}                  & 2.24                          & 81.46                        & 1826                        & 1826                        \\
RDN\cite{zhang2020rdnir}                   & 0.53                          & 26.63                        & 526                         & 584                         \\
MIRNet\cite{Zamir2020MIRNet}                & 7.95                          & 92.26                        & 1735                        & 1956                        \\
FBCNN\cite{jiang2021towards}                & 71.92                          & 134.74                        & 3021                        & 3597                        \\
JARCAL\cite{wang2022jpeg}                & 13.88                          & 419.45                        & 7281                        & 8345                        \\
DDP\cite{wang2020deep}                & 0.14                          & 1.48                        & 102                        & 98                        \\ \hline
Our                   & \textbf{0.11}                          & \textbf{0.36}                         & \textbf{54}                          & \textbf{58}                          \\ \bottomrule
\end{tabular}
\end{table}

\subsection{Qualitative Comparison}
We present the degraded features and feature de-drifting results of different methods in Fig. \ref{fig1-CompareWithOtherMethods}. Since JPEG compression applies DCT on each block, the correlation between adjacent blocks is ignored, which introduces spatial-varied feature drifting. Besides, the amplitude of the feature response in discriminative regions is weakened due to the loss of high-frequency components. The existing JAR methods can effectively improve human visual perception, but they cannot significantly eliminate spatial-varied feature drifting, as shown in Fig. \ref{fig1-CompareWithOtherMethods}. The reason for this is that the enhanced results of these JAR methods tend to be over-smoothed, and some critical details are lost, which may not be helpful for CNN-based image classification. Compared with these JAR methods, our method can effectively correct the spatial-varied feature drifting and improve the feature response in the discriminative region.

Next, we compare the residual maps estimated by FE-Net with or without the guidance of FDE-Net as shown in Fig. \ref{fig:fig-ResidualComparison}. We can observe that the estimated residual under the guidance of FDM is closer to the real residual distribution and the corresponding feature enhancement result in Fig. \ref{fig:fig-ResidualComparison} (g) has more complete structure information. This demonstrates that our FDE-Net can accurately estimate the amplitude of feature drifting for each block. Under the guidance of FDM, the feature response on discriminative regions can be effectively recovered. Moreover, we also visualize the feature enhancement results of our method and JAR methods in multiple compression scenarios, as shown in Fig. \ref{fig:fig1-multi-compressed}. We can see that as the compression time T increases, the feature drifting will become more serious, and the JAR methods cannot guarantee the enhancement of effective features. The reason for this is that image enhancement methods mainly pursue to obtain visually pleasing effects and cannot achieve satisfactory performance on downstream high-level tasks \cite{yang2023visual,son2020urie,pei2018does,DDP} and it cannot guarantee that the regions with similar structures in the image can be enhanced uniformly, leading to uncovered and incomplete feature representation for classification \cite{yang2023visual,son2020urie,pei2018does}. For example, for the polar bear in Fig. \ref{fig:fig1-multi-compressed}, the JAR methods can hardly enhance the polar bear structure feature. However, our method can better enhance the structural feature of polar bears, which is much closer to the clear feature. This demonstrates that, compared to JAR methods, our method can better enhance the object features in the pre-trained model under various compression conditions.

\begin{figure}[h]  
    \centering
    
    \includegraphics[width=0.9\linewidth]{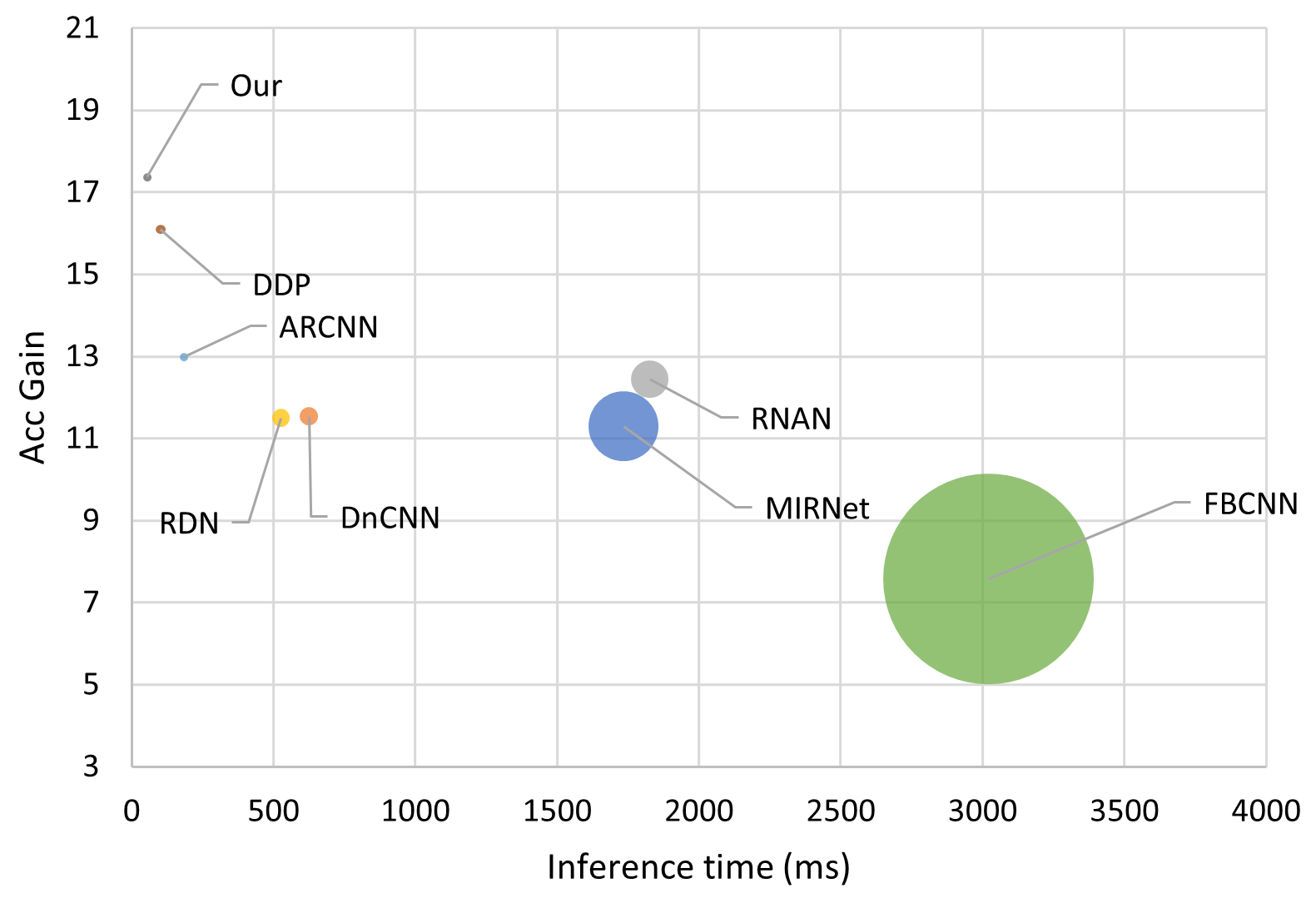}
    \caption{The comparison of inference time and recognition performance gain on mobile CPU. The experiment is conducted on ResNet-50 under QF=15. Note that the size of points represents the parameters of models.}
    \label{fig:infer}
    
\end{figure}

\begin{figure}[tbp]  
    \centering
    \includegraphics[width=0.9\linewidth]{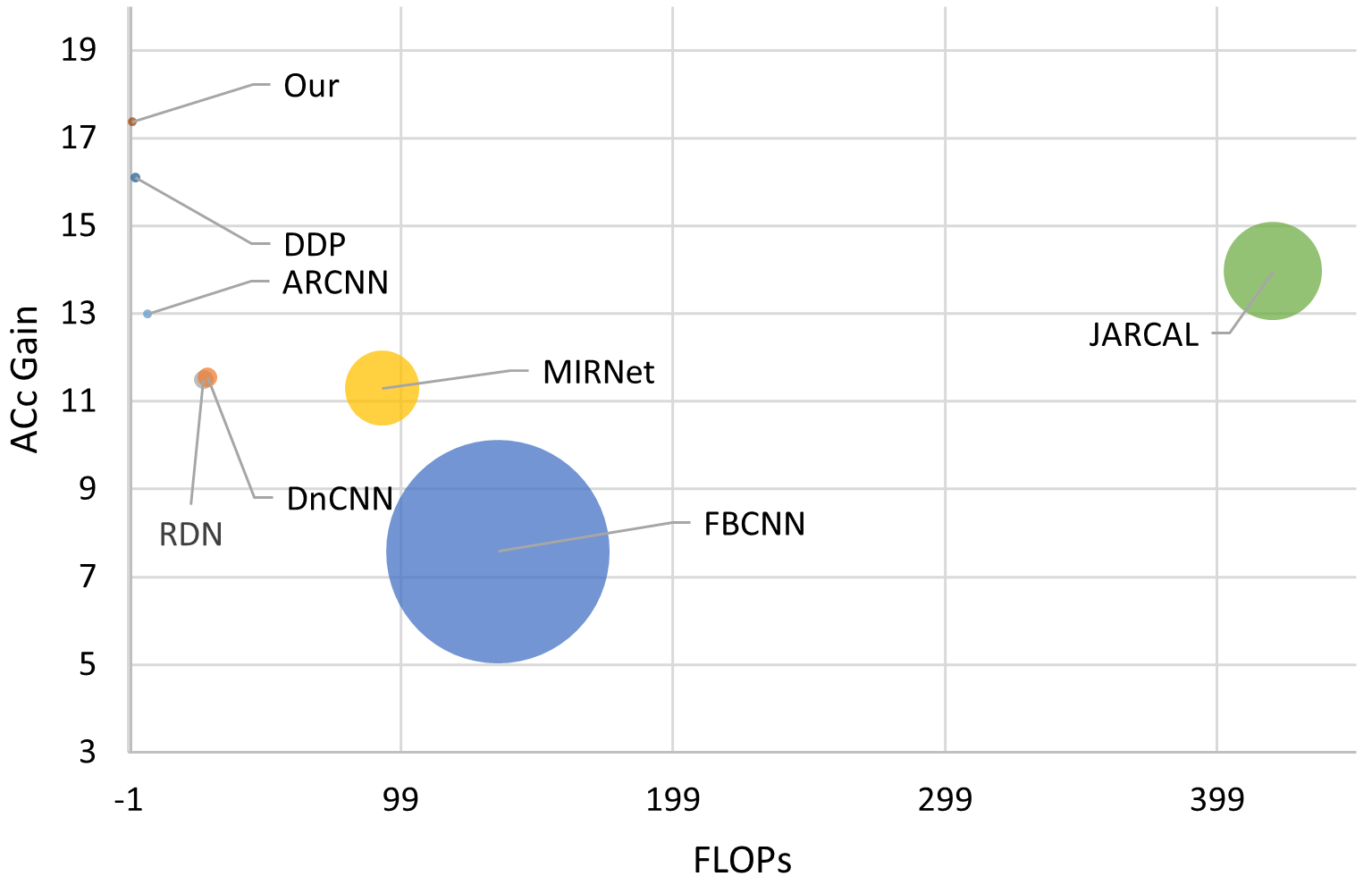}
    \caption{The comparison of FLOPs and recognition performance gain on mobile CPU. The experiment is conducted on ResNet-50 under QF=15. Note that the size of points represents the parameters of models.}
    \label{fig:flops}    
\end{figure}

\subsection{Performance Analysis}
\textbf{1. Efficiency evaluation. \quad}The efficiency of the model is a crucial part of mobile device deployment, which determines whether it can be widely used in mobile devices. We conduct inference time comparisons on mobile devices (Snapdragon 865) to explore the deployment potential. Note that the CPU tend to be specifically optimized on mobile devices, which may cause the situation that inference speed on the CPU is faster than on GPU. As shown in Fig. \ref{fig:infer}, our method is more than 3 times faster than those JAR methods on both GPU and CPU. The FLOPs of our method are 10 to 100 times smaller than that of JAR methods thanks to the lightweight model architecture and structural re-parameterizatio, as shown in Fig.\ref{fig:flops}. Besides, our method requires the least number of parameters compared to other methods, which demonstrates we have the lowest memory requirements, as shown in Table \ref{tab:inf on mobile}.

\begin{figure}[tbp]  
    \centering
    
    \includegraphics[width=0.9\linewidth]{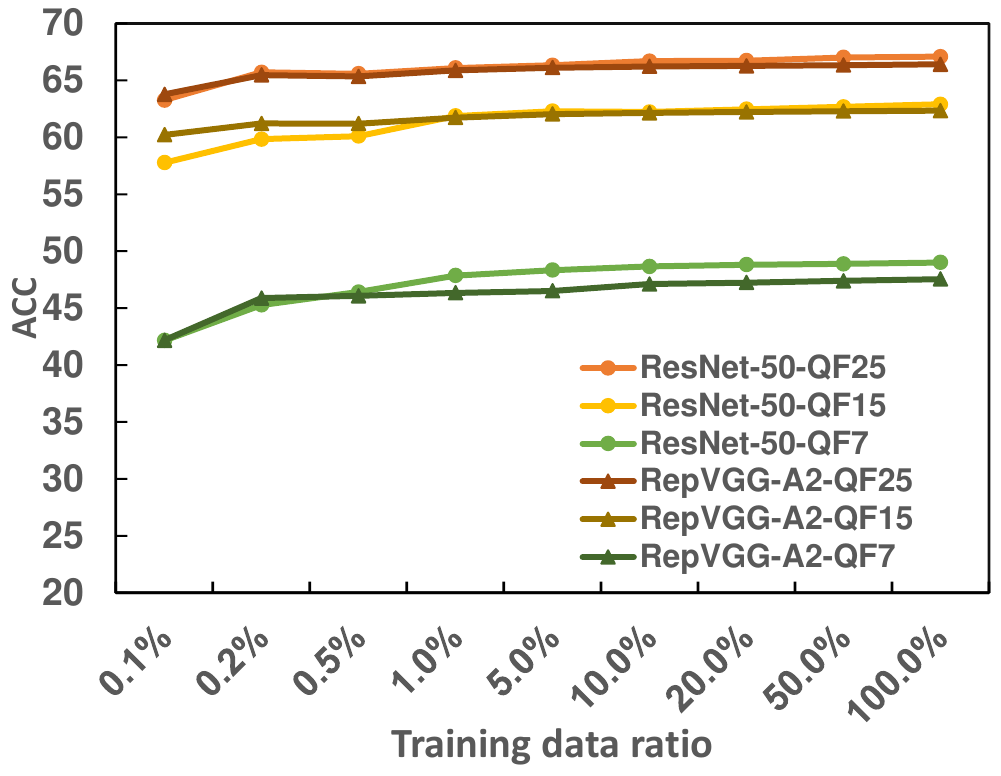}
    \caption{The performance of the different sizes of training data on pre-trained classification models under various compression conditions.}
    \label{fig:trainingraatio}
    
\end{figure}

\textbf{2. The size of training data. \quad}Considering that there are often difficulties in collecting datasets in real scenarios, we analyze the impact of the performance of the proposed method on the size of the training dataset. We conduct nine different sizes of training data and evaluate the classification performance under three different compressed conditions. As shown in Fig.\ref{fig:trainingraatio}, our proposed method can still achieve remarkable classification improvement in both ResNet-50 and RepVGG-A2 with limited training data. The pre-trained model can still obtain at least 10$\%$ accuracy improvement even only trained with 20 images (~$0.1\%$ of total 20000 training images), demonstrating that our method is less dependent on the size of training data and able to adapt to data-scarce scenarios.

\begin{table}[ht]
\centering
  \caption{Comparison with the task-driven method.}
\begin{tabular}{c||ccccc}
\hline
ResNet-50  & 25             & 18            & 15             & 10             & 7              \\ \hline
DirectTest & 62.59          & 59.46         & 54.82          & 41.59          & 26.22          \\
URIE       & 63.60          & 62.17         & 58.09          & 52.24          & 42.80          \\
Our        & \textbf{68.21} & \textbf{64.30} & \textbf{62.19} & \textbf{54.64} & \textbf{45.34}  \\
\hline
\end{tabular}
\label{tab:URIE}
\end{table}

\textbf{3. Comparison with the task-driven method. \quad} We also compare our methods with the task-driven method URIE~\cite{son2020urie}, which is supervised by semantic labels to obtain a task-friendly enhancement model. For a fair comparison, we randomly select 200 images from the CUB-200\cite{wah2011caltech} as supervised labels and use DnCNN~\cite{zhang2017beyond_DNCNN} as the JAR model for URIE~\cite{son2020urie} on ResNet-50. The results are shown in Table \ref{tab:URIE}. After training, URIE can slightly improve the classification performance, compared to inferencing directly from the degraded images. Compared with URIE, our method can achieve better classification performance, especially under heavy compression conditions.
\begin{table}[h]
\centering
\scriptsize

\caption{The image classification performance for imagenet dataset on AVIF image format with quality factors (QF) 25, 18, 15, 10, and 7. Avg means the average classification performance of different QF. The best and the second-best classification accuracy (\%) are marked as \textbf{bold} and \underline{underline}.}
\begin{tabular}{c|cccccc}
\hline
AVIF                       & QF  & Direct Test & PANet & Restormer & DDP  & Ours  \\ \hline
\multirow{6}{*}{ResNet-50} & 25  & 64.18       & 63.03           & 64.49             & \underline{64.51} & \textbf{65.67} \\
                           & 18  & 60.51       & 61.54           & 62.25             & \underline{62.64} &  \textbf{63.79} \\
                           & 15  & 57.38       & 58.22           & 59.36             & \underline{59.67} &  \textbf{60.87} \\
                           & 10  & 53.09       & 55.32           & \underline{56.44}             & 56.28 &  \textbf{57.93} \\
                           & 7   & 47.42       & 51.86           & \underline{52.75}             & 52.12 &  \textbf{53.32} \\
                           & Avg & 56.51       & 57.99           & \underline{59.06}             & 59.04 &  \textbf{60.32} \\ \hline
\end{tabular}
\label{table:AVIF}
\vspace{-5mm}
\end{table}

\begin{table}[h]
\centering
\scriptsize
\caption{Comparsion with transfer learning. The best and the second-best classification accuracy (\%) are marked as \textbf{bold} and \underline{underline}.}

\begin{tabular}{c|cccc}

\hline
                           & QF                             & Direct Test & Transfer learning & Ours  \\ \hline
\multirow{6}{*}{ResNet-50} & 25                             & 66.16       & \underline{66.17}             & \textbf{67.10} \\
                           & 18                             & 62.47       & \underline{64.25}             & \textbf{64.79} \\
                           & 15                             & 59.28       & \underline{62.12}              &\textbf{62.91} \\
                           & 10                             & 47.50       & \underline{56.61}              &\textbf{57.24} \\
                           & 7                              & 31.65       & \textbf{50.40}             & \underline{49.02} \\
                           & clear images & 76.04       & 70.32             & \textbf{76.04} \\ \hline
\end{tabular}
\label{table:Transfer}
\vspace{-5mm}
\end{table}

\textbf{4. Generalization for other lossy compression formats.\quad}To evaluate the performance of the proposed method for other lossy compression formats, AV1 Image File Format (AVIF) \cite{AVIF} is evaluated, and the image quality factor (QF) of the validation set of imagenet \cite{Imagenet} is set to 25, 18, 15, 10, and 7, respectively. Due to the lack of dedicated image restoration algorithms for AVIF compression enhancement, we choose the general image restoration networks PANet (IJCV,2023)\cite{PANet} and Restormer(CVPR,2022) \cite{zamir2022restormer} for comparison, as well as the classic feature enhancement method DDP (CVPR,2020)\cite{DDP}, and the results are shown in Table \ref{table:AVIF}. It can be observed that as the QF decreases, the recognition performance of the pre-trained image classification model gradually decreases. This is because as the image compression degree increases and the image quality decreases, image details and textures are increasingly lost, causing feature drifting in the features of the pre-trained model \cite{DDP}, thereby damaging image recognition performance. Compared with the above methods, the proposed AFD-Module can more significantly improve the recognition performance of pre-trained models at different compression rates. For example, for AVIF compressed images with QF=7, directly using the pre-trained Resnet-50 \cite{resnet} for recognition has a performance of 47.42\%. After introducing the proposed AFD-Module, the performance of the Resnet-50 is improved by 5.9\%, which is 1.46\%, 0.57\%, and 1.20\% higher than PANet, Restormer, and DDP methods, respectively. This proves the effectiveness and versatility of the proposed method for AVIF lossy compression format.

\textbf{5. Comparison with transfer learning.\quad}
To demonstrate the superiority of our proposed method, we compare our method with transfer learning. Specifically, we performed transfer learning on a pre-trained ResNet-50 network under JPEG scenarios. The results are shown in the Table \ref{table:Transfer}. It can be seen that after fine-tuning, ResNet-50 improves image classification performance in the target domain (QF is 7-25) but inevitably suffers performance degradation in the source domain. For example, after fine-tuning, ResNet-50 can improve image classification performance by 9.11\% under a compression scenario with QF of 10 but will drop the source domain and reduce network classification performance in natural scenes by -5.72\%. However, our method can achieve better image classification performance improvement under different compression scenarios while not affecting the source domain, that is, maintaining network classification performance for clean images. This demonstrates our method’s flexibility and transferability.


\section{Conclusion}
In this paper, we propose a novel adaptive feature de-drifting module for compressed image classification. A feature drifting estimation network is devised to generate the feature drifting map by exploring the frequency distribution within each block, which is utilized to guide the feature de-drifting. After training on limited images without semantic supervision, the module can be directly plugged into the pre-trained models to boost the classification performance on compressed images.

\ifCLASSOPTIONcaptionsoff
  \newpage
\fi
\bibliographystyle{IEEEtran}
\bibliography{IEEEabrv,main}
\end{document}